\documentclass[a4paper,fleqn]{cas-dc}

\usepackage[authoryear]{natbib}

\usepackage[dvipsnames]{xcolor}

\newcommand{\BO}[1]{{\boldsymbol{#1}}}

\newcommand{\OP}[1]{{\operatorname{#1}}}
\usepackage{booktabs}       
\usepackage{threeparttable} 
\usepackage{graphicx}       
\usepackage{colortbl}
\usepackage{xcolor}

\def\tsc#1{\csdef{#1}{\textsc{\lowercase{#1}}\xspace}}
\tsc{WGM}
\tsc{QE}

\begin{document}
\let\WriteBookmarks\relax
\def\floatpagepagefraction{1}
\def\textpagefraction{.001}

\shorttitle{}

\shortauthors{}

\title [mode = title]{Expert Knowledge-Guided Decision Calibration for Accurate Fine-Grained Tree Species Classification}



%

\author[1]{Chen Long}[orcid=0000-0001-7473-6852]
\ead{chenlong107@whu.edu.cn}

\author[1]{Dian Chen}
\ead{chendian@whu.edu.cn}

\author[1]{Ruifei Ding}
\ead{dingrf@whu.edu.cn}

\author[1]{Zhe Chen}
\ead{ChenZhe_WHU@whu.edu.cn}

\author[1,2]{Zhen Dong}[orcid=0000-0002-0152-3300]
\ead{dongzhenwhu@whu.edu.cn}
\cormark[1]

\author[1,2]{Bisheng Yang}[orcid=0000-0001-7736-0803]
\ead{bshyang@whu.edu.cn}

\affiliation[1]{organization={State Key Laboratory of Information Engineering in Surveying, Mapping and Remote Sensing, Wuhan University},
            city={Wuhan},
            state={430079},
            country={China}}
\affiliation[2]{organization={Hubei Luojia Laboratory},
            city={Wuhan},
            state={430079},
            country={China}}
\cortext[1]{Corresponding author}



\begin{abstract}
Accurate fine-grained tree species classification is critical for forest inventory and biodiversity monitoring. 
Existing methods predominantly focus on designing complex architectures to fit local data distributions. However, they often overlook the long-tailed distributions and high inter-class similarity inherent in limited data, thereby struggling to distinguish between few-shot or confusing categories.
In the process of knowledge dissemination in the human world, individuals will actively seek expert assistance to transcend the limitations of local thinking.
Inspired by this, we introduce an external "Domain Expert" and propose an Expert Knowledge-Guided Classification Decision Calibration Network (EKDC-Net) to overcome these challenges.
Our framework addresses two core issues: expert knowledge extraction and utilization. 
Specifically, we first develop a Local Prior Guided Knowledge Extraction Module (LPKEM). By leveraging Class Activation Map (CAM) analysis, LPKEM guides the domain expert to focus exclusively on discriminative features essential for classification. Subsequently, to effectively integrate this knowledge, we design an Uncertainty-Guided Decision Calibration Module (UDCM). This module dynamically corrects the local model's decisions by considering both overall category uncertainty and instance-level prediction uncertainty.
Furthermore, we present a large-scale classification dataset covering 102 tree species, named CU-Tree102 to address the issue of scarce diversity in current benchmarks. 
Experiments on three benchmark datasets demonstrate that our approach achieves state-of-the-art performance. Crucially, as a lightweight plug-and-play module, EKDC-Net improves backbone accuracy by 6.42\% and precision by 11.46\% using only 0.08M additional learnable parameters.
The dataset, code, and pre-trained models are available at \url{https://github.com/WHU-USI3DV/TreeCLS}.
\end{abstract}



\begin{keywords}
Expert Knowledge Fusion \sep Deep Learning \sep Fine-grained Tree classification
\end{keywords}

\maketitle

\section{Introduction}
\label{sec:intro}

\begin{figure*}[t]
    \centering
    \includegraphics[width=1.0\textwidth]{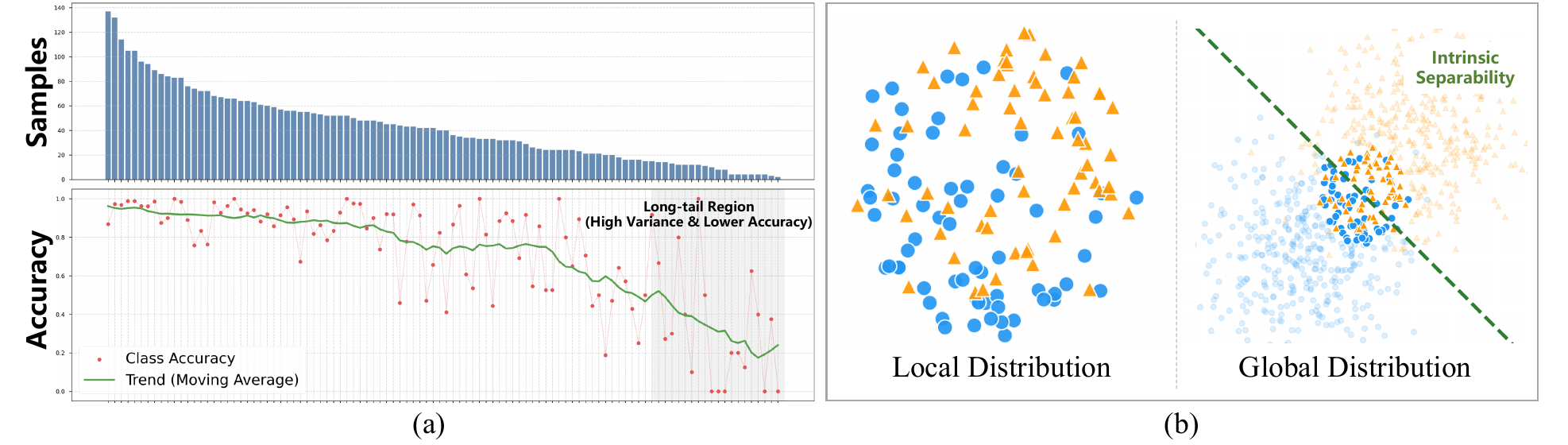}
    \caption{Illustration of the limitations in existing methods. (a) Struggle with Long-tail Categories: The upper histogram illustrates the imbalanced sample distribution, while the lower plot reveals the corresponding drop in classification accuracy. (b) Ambiguity in Local Distributions: Although the global distribution exhibits inherent separability, the local distribution reveals significant overlap and ambiguity along the decision boundary, due to the limitations of the data.}
    \label{fig:teaser}
\end{figure*}

Trees are vital components of ecosystems, playing an indispensable role in protecting biodiversity \citep{barrios2018contribution}, promoting the sustainable development of the environment \citep{honeck2018vegetation}, and regulating the global carbon cycle \citep{dong2025remote}. The efficient and accurate fine-grained classification of trees serves as a fundamental prerequisite for facilitating precision forest management \citep{laumer2020geocoding} and advancing green city initiatives \citep{chen2023developing}.

Traditional field survey methods are labor-intensive, time-consuming, and difficult to scale \citep{hansen1996classification}. With the rapid advancement of earth observation methods and computer vision technologies \citep{dong2025neural}, image-based deep learning approaches have emerged as powerful tools for rapid tree species identification \citep{kwon2023merging}. \cite{carpentier2018tree} constructed the first bark image dataset and achieved accurate classification by combining convolutional neural networks (CNN) with majority voting strategy. \cite{rzanny2019flowers} integrated information from multiple organs (e.g., flowers and leaves) as an effective method to improve classification accuracy. Recently, the emergence of large-scale plant datasets like iNat\footnote{\url{https://www.inaturalist.org/}} and Pl@ntNet-300K \citep{garcin2021pl} has spurred increasing research interest in developing complex fine-grained visual classification algorithms to capture the complex morphological subtle differences between different tree species \citep{branson2018google,dong2025multiscale}.

Although these methods have achieved remarkable results through intricate designs, applying them to fine-grained tree species classification remains challenging.
A fundamental limitation is that \textit{they only focus on fitting the fixed distribution of the local training dataset, while ignoring the difference between this restricted local view and the real data distribution, resulting in the learning of biased decision boundaries}. This manifests in two major drawbacks:
1) \textbf{Struggle with Long-tail Categories:} As shown in Fig.\ref{fig:teaser}-(a), due to the high cost of expert labeling and data collection, datasets often suffer from severe class imbalance. Models fitting such local distributions fail to generalize to tail classes, leading to poor performance and high variance. 2) \textbf{Ambiguity in Local Distributions:} In local views, feature distributions of different species may overlap due to atypical samples (e.g., two similar poplar and locust trees). Focusing too much on these local details causes the model to overlook the intrinsic separability of the global data. Consequently, the model tends to learn biased decision boundaries, making it difficult to distinguish confusing categories (see Fig.\ref{fig:teaser}-(b)).

To address this challenge, we draw inspiration from the process of knowledge transmission in human cognition. In human learning, when a student is confused by ambiguous concepts or limited examples, they typically seek guidance from a knowledgeable teacher possessing a broader, global perspective. Motivated by this process, we propose a novel Expert Knowledge-Guided Classification Decision Calibration Network, named EKDC-Net. Instead of relying solely on the limited features, it introduces an external domain expert to rectify the classification decisions of the local backbone.
Specifically, we employ advanced vision foundation models \citep{stevens2024bioclip,gu2025bioclip} to serve as domain experts. Benefiting from contrastive pre-training \citep{clip} on massive paired biological image-text data, these models offer robust global representations with intrinsic separability, effectively compensating for the limited views of local models.
To effectively \textbf{extract} and \textbf{utilize} this expert knowledge, we design two specialized modules. First, we propose the \textbf{Local Prior-Guided Knowledge Extraction Module (LPKEM)}. By leveraging Class Activation Map (CAM) \citep{cam} analysis in the local model as spatial cues,  LPKEM filters out irrelevant background noise, guiding the domain expert to focus strictly on discriminative core knowledge (e.g., foliage, bark). Subsequently, bypassing complex knowledge distillation \citep{hu2025tedfl} or feature fusion mechanisms \citep{long2024sparsedc}, we design the \textbf{Uncertainty-Guided Decision Calibration Module (UDCM)} to perform logit-level decision correction. It dynamically calibrates the local backbone's decisions by jointly analyzing global category-level uncertainty and instance-level prediction uncertainty, thereby correcting misclassifications stemming from limited local views. Furthermore, to address the lack of diversity in current benchmarks \citep{yang2023urban}, we construct and release a large-scale classification dataset covering 102 tree species in China. This provides a challenging benchmark that reflects real-world complexity.

In summary, the main contributions of this work are as follows:
\begin{itemize}
    \item We analyze the limitations of fitting local distributions and propose EKDC-Net, a plug-and-play framework to address the challenges of long-tailed and confusing classes. Specifically, we design LPKEM to extract discriminative knowledge using local priors, and UDCM to calibrate decisions based on uncertainty estimation, thus improving classification accuracy.
    \item We introduce a new large-scale dataset comprising 102 tree species, named CU-Tree102. It significantly expands the diversity and scale of existing datasets, providing a more challenging test benchmark for fine-grained image classification tasks in forestry.
    \item Extensive experiments demonstrate that the proposed EKDC-Net achieves state-of-the-art performance. With only \textbf{0.08M} additional learnable parameters, it improves the backbone accuracy by \textbf{6.42\% }and precision by \textbf{11.46\%}, and enhances the average generalization capability across different domains by \textbf{30.56\%}.
\end{itemize}
\section{Related Work}
\label{sec:related}
\begin{table*}[t]
\centering
\caption{Comparison with existing image-based tree species classification datasets.}
\label{tab:dataset_comparison}
\begin{tabular}{lcccccc}
\toprule
\textbf{Dataset} & \textbf{Part} & \textbf{Taxonomic} & \textbf{Classes} & \textbf{Images} & \textbf{Region} & \textbf{Access} \\ \midrule
Swedish Leaf \citep{soderkvist2001computer} & Leaf & Species & 15 & 1,125 & Sweden & Public \\
Flavia \citep{wu2007leaf} & Leaf & Species & 32 & 1,907 & China & Public \\
LeafSnap \citep{kumar2012leafsnap} & Leaf & Species & 185 & 30,866 & USA & Public \\
BarkNet 1.0 \citep{barrios2018contribution} & Bark & Species & 23 & 23,000 & Canada & Public \\ 
BarkVN-50 \citep{gbt4tdmttn-1} & Bark & Species & 50 & 5,768 & Vietnam & Public \\ \midrule
Pasadena Urban \citep{wegner2016cataloging} & Whole-plant & Species & 18 & 80,000 & USA & Restricted \\
Auto Arborist \citep{beery2022auto} & Whole-plant & Genus & 300 & 2.5M & N. America & Restricted \\
Jekyll \citep{yang2023urban} & Whole-plant & Species & 23 & 4,804 & China & Public \\ \midrule
\textbf{CU-Tree102 (Ours)} & \textbf{Whole-plant} & \textbf{Species} & \textbf{102} & \textbf{9,134} & \textbf{China} & \textbf{Public} \\ \bottomrule
\end{tabular}
\end{table*}

\subsection{Fine-Grained Tree Species Classification}
Fine-Grained tree species Visual Classification (FGVC) is a critical task in urban forestry, aiming to distinguish subordinate categories with subtle morphological differences \citep{arevalo2024challenges}.

Early approaches primarily focused on applying standard backbone networks to extract high-level semantic representations. Architectures have evolved rapidly from Convolutional Neural Networks (CNNs) \citep{chou2022novel,lee2023mapping} to Vision Transformers \citep{sun2022sim}, ultimately dominated by  Swin Transformers \citep{chen2024fet,wu2024swin} with multi-scale features and global modeling capabilities.
However, due to the high intra-class variance of trees (e.g., appearance changes across different seasons) and subtle inter-class similarities, these generic backbones often struggle to capture the critical details required for accurate identification.

To address these limitations, researchers have adapted general Fine-Grained Visual Classification techniques to forestry. The core idea is to identify discriminative local parts between tree species, such as the unique venation of leaves or the texture of the bark.
Following this idea, some methods \citep{chou2023fine,chou2022novel} use an auxiliary branch to generate binary importance masks, which filter out irrelevant backgrounds to focus only on core features. Other works \citep{song2022non} utilize attention mechanisms to locate key components and employ heatmap pseudo-labels from Class Activation Maps (CAMs) for supervision, ensuring the model captures the correct classification cues \citep{wang2024multi}.
Specific to tree species, \cite{dong2025multiscale} proposed a multi-scale feature fusion Transformer that integrates morphological features at different scales to capture complex structures such as bark, fissures, and so on. To further enhance discriminability beyond appearance, some methods attempt to capture hierarchical biological concepts \citep{liu2022focus,wang2023consistency} or high-level semantic features \citep{cgl} to distinguish subtle differences between species.

However, despite these complex designs that have helped them achieve superior performance on benchmark datasets, a fundamental limitation remains: these methods mainly optimize their attention maps or concept bases to fit a fixed distribution of local training data, ignoring the inherent gap between the limited training samples and the global data distribution. As a result, they still struggle with the long-tailed distribution and high inter-class similarity inherent in real world.

\subsection{Vision Foundation Models as Experts}
To overcome the data scarcity in specific domains, leveraging Vision Foundation Models (e.g., CLIP \citep{clip}, DINOv3 \citep{simeoni2025dinov3}) has become a promising direction. These models capture a wide range of open-world knowledge from large-scale datasets through self-supervised pre-training. They can act as domain experts with a global perspective to improve the performance of downstream tasks, such as image segmentation \citep{hu2025tedfl}, image detection \citep{liu2024grounding}, 3D reconstruction \citep{long2024sparsedc}, etc.

Current paradigms generally fall into two categories: (1) Knowledge Distillation. This approach focuses on aligning the student network's internal representations with the expert's latent space. For example, \cite{hu2025tedfl} introduced auxiliary branching of characteristics to require the visual output of the backbone to remain consistent with the high-dimensional embeddings of CLIP. (2) Feature Fusion. This approach typically employs heavy architectural stacking, such as multiple layers of cross-attention, to facilitate the fusion of local visual features with expert embeddings \citep{zhang2025region}. However, these methodologies encounter significant bottlenecks in the field of fine-grained tree species classification. On one hand, they often require vast amounts of paired data to bridge the representational gap between the foundation model and the task-specific backbone. On the other hand, the introduction of numerous learnable fusion layers substantially increases the risk of overfitting, especially when training on limited specialized samples.

Compared to existing solutions, our method avoids the need for extensive cross-modal alignment or deep feature distillation. Tailored to the intrinsic nature of classification tasks, we utilize the expert's knowledge solely to calibrate the classification decisions (i.e., logits) of the backbone. This lightweight, plug-and-play design effectively addresses the mentioned challenges.

\subsection{Tree Species Classification Datasets}
The development of fine-grained classification methods is highly dependent on the quality of available datasets.
In the professional field of tree species classification, existing benchmarks can be divided into organ-centric (e.g., leaves and bark) and whole-plant datasets.

\textbf{Organ-centric Datasets.} Due to the stable morphological characteristics of leaves, earlier research focused on leaf-based identification. Datasets such as Swedish Leaf \citep{soderkvist2001computer}, Flavia \citep{wu2007leaf}, and LeafSnap \citep{kumar2012leafsnap} provided high-quality leaf scan images, establishing the initial benchmarks for the field.
To address the seasonal limitations of leaf-based tree species classification (i.e., some deciduous trees do not have leaves in winter), researchers have introduced bark-centric datasets such as BarkNet 1.0 \citep{barrios2018contribution} and BarkVPN-50 \citep{gbt4tdmttn-1}. These works utilize the stable texture feature of tree bark to ensure consistent classification across seasons.
However, a significant gap still exists between these organ-centric methods and real-world applications. Most images have clean backgrounds and good lighting, which fail to capture the complexity of forestry or urban environments in the real world. More importantly, capturing such fine-grained details (e.g., complex venation of leaves, the texture of bark cracks) is usually impractical in large-scale forestry surveys \citep{arevalo2024challenges}.

\textbf{Whole-plant Datasets.} To bridge this gap, recent research has begun to use easily accessible crowdsourced images to create datasets containing entire trees. Pasadena Urban Trees \citep{wegner2016cataloging} and the Auto Arborist Dataset \citep{beery2022auto} are pioneering works in this category. The former includes 80,000 trees in California, whereas the latter covers 2.5 million trees over 300 genera across North America.
Despite their large scale, these datasets are difficult to access due to data privacy policies and regional restrictions.
Among the few open-source alternatives, the Jekyll dataset \citep{yang2023urban} provides 4,804 images of more than 23 species in multiple cities across China. However, the lack of diversity in these images (i.e., represented by a simple backbone network with over 95\% accuracy) makes it insufficient for evaluating advanced fine-grained algorithms.

To address these challenges, we propose \textbf{CU-Tree102}, a large-scale tree species dataset containing 9,134 images across 102 species. CU-Tree102 meets the urgent demand for an accessible and diverse benchmark in this field, capturing trees within complex, real-world backgrounds. Tab.\ref{tab:dataset_comparison} provides a comprehensive comparison between our dataset and existing benchmarks.

\section{Methodology}
\label{sec:methodology}

\begin{figure*}[t]
    \centering
    \includegraphics[width=1.0\linewidth]{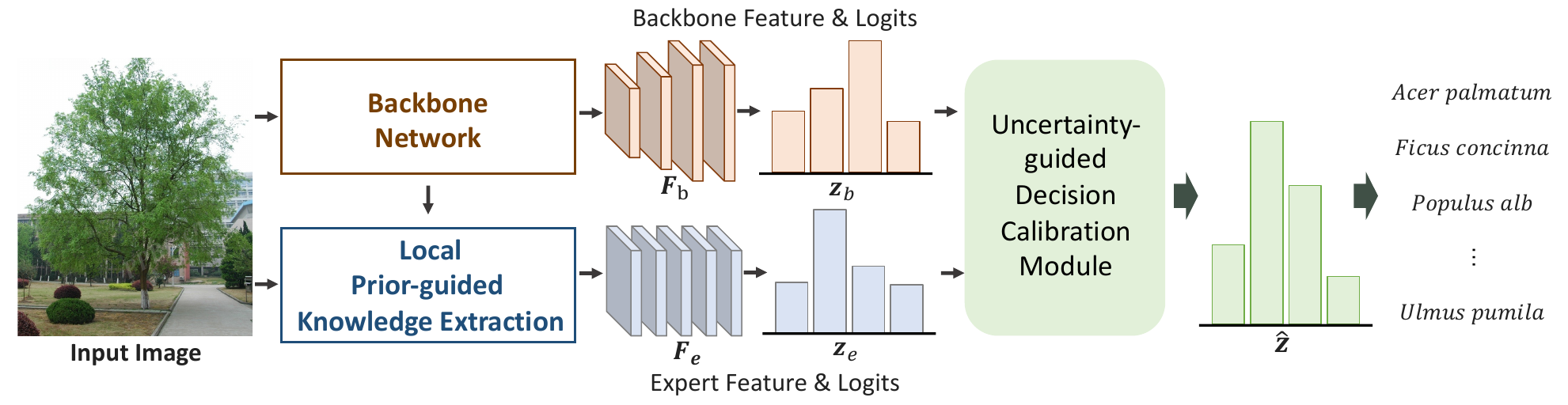}
    \caption{Overview of EKDC-Net architecture. Given a tree image, EKDC-Net accurately identifies the fine-grained tree species.}
    \label{fig:overview}
\end{figure*}

\subsection{Overview}

Given an optical image $\mathcal{I} \in \mathbb{R}^{H \times W \times 3}$ containing trees, our goal is to accurately classify the corresponding tree species. Existing methods focus on fitting local data distributions, which leads to struggles with the inherent challenges of long-tail distributions and high inter-class similarities in forestry. Unlike these methods, we propose an Expert Knowledge-guided Decision Calibration Network (EKDC-Net). The core idea of EKDC-Net is to introduce a visual foundation model as an external "domain expert" and utilize its extensive knowledge gained through massive pre-training to overcome the limitations of local data distributions.

As shown in Figure \ref{fig:overview}, our proposed framework consists of a backbone network and two lightweight modules: the Local Prior-guided Knowledge Extraction Module (LPKEM) and the Uncertainty-guided Decision Calibration Module (UDCM). First, the backbone network processes the image $\mathcal{I}$ to extract multi-scale features $\BO{F}_b = \{\BO{f}_b^1, \BO{f}_b^2, \BO{f}_b^3, \BO{f}_b^4\}$ and produce the initial classification logits $\BO{z}_b \in \mathbb{R}^K$, where $K$ is the number of categories. Next, the LPKEM takes these features, the original image, and the initial logits to generate expert features $\BO{F}_e = \{\BO{f}_e^0, \BO{f}_e^1, \BO{f}_e^2, \BO{f}_e^3, \BO{f}_e^4\}$ and corresponding logits $\BO{z}_e \in \mathbb{R}^K$. Finally, UDCM takes $\BO{z}_b$ and $\BO{z}_e$ as inputs to dynamically and stably correct the decisions of the backbone, thus achieving accurate classification.

\subsection{Local Prior-guided Knowledge Extraction}
The primary goal of LPKEM is to extract robust taxonomic knowledge from the domain expert.
However, due to factors such as collection conditions and environment, trees often become visually entangled with complex backgrounds.
Therefore, indiscriminately feeding the entire image into a frozen expert model introduces significant noise, leading to suboptimal feature extraction.

\begin{figure*}[t]
    \centering
    \includegraphics[width=1.0\linewidth]{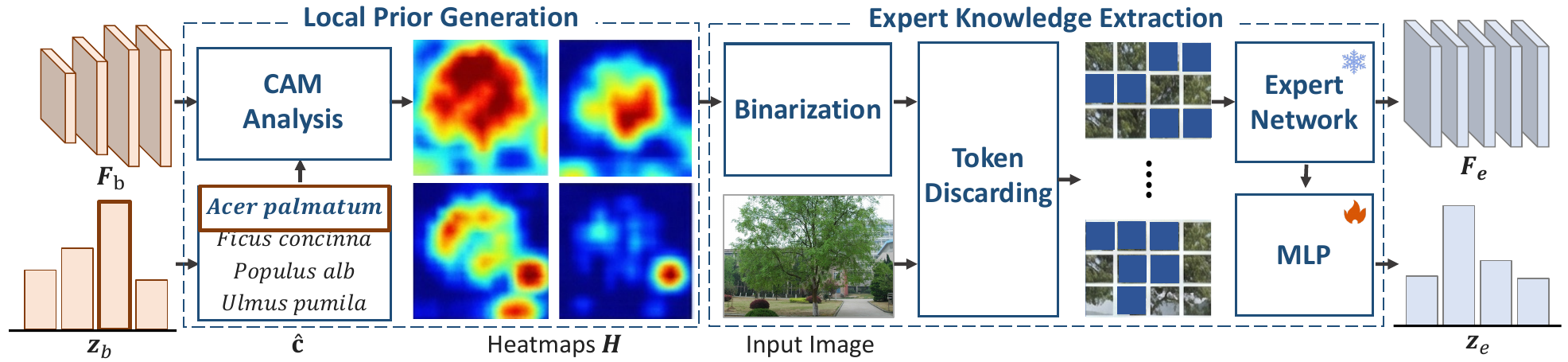}
    \caption{An illustration of the Local Prior-guided Knowledge Extraction Module. \includegraphics[width=1.2em]{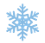} means frozen this network parameters, \includegraphics[width=1.2em]{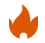} denotes the trainable network parameters.}
    \label{fig:lpkem}
\end{figure*}

We draw inspiration from existing mechanisms of academic inquiry: when students turn to their tutor for guidance, they don't ask the tutor to solve the problem directly, but pinpoint the specific cause of the confusion. This targeted context allows the tutor to directly address the core issue without sifting through extraneous information. Similarly, while the local backbone may misclassify fine-grained species, it generally retains a local experience of which parts (e.g., leaves or texture of the bark) of the picture are important for classification.
Inspired by this, we design LPKEM as shown in Fig.\ref{fig:lpkem} to extract expert knowledge. It extracts spatial cues from the local model to direct external experts to focus on discriminant regions.

\textbf{Local Prior Generation.}
Gradient-weighted Class Activation Map Analysis \citep{cam} is a widely established technique for identifying discriminative regions that contribute most significantly to model decisions. We regard these regions as the core spatial cues for guiding the external expert. However, standard CAM computation relies on ground-truth labels to calculate gradients, which are unavailable during the inference phase.
To solve this problem, we take the category index with the largest response in the local logits $\BO{z}_b$ as the pseudo-label for classification, denoted as $\hat{c}=\OP{argmax}(\BO{z}_b)$. Based on this pseudo-label, we generate a set of hierarchical heatmaps $\BO{H} = \{\BO{h}_1, \BO{h}_2, \BO{h}_3, \BO{h}_4\}$ corresponding to the backbone's multi-scale feature maps.
Specifically, each heatmap $\BO{h}_l \in \mathbb{R}^{H_l \times W_l}$ spatially aligns with the feature map $\BO{f}_b^l \in \mathbb{R}^{C_l \times H_l \times W_l}$ at the $l$-th stage, where $C_l$ denotes the number of channels. It follows a two-step process. First, we use Eq.\ref{eq:cam_weight} to quantify the importance of each feature channel.
\begin{equation}
\label{eq:cam_weight}
    \alpha_k^l = \operatorname{GAP}\left( \frac{\partial y^{\hat{c}}}{\partial \BO{f}_b^{l,k}} \right).
\end{equation}
In this step, $\alpha_k^l$ represents the importance of the $k$-th feature channel of the $l$-layer feature, the $\operatorname{GAP}(\cdot)$ denotes the Global Average Pooling operation. The term $\frac{\partial y^{\hat{c}}}{\partial \BO{f}_b^{l,k}}$ represents the gradient map, quantifying the sensitivity of the classification predictions with respect to the $k$-th feature map channel $\BO{f}_b^{l,k}$. Here, $y^{\hat{c}}$ refers to the raw logit score of the predicted pseudo-label $\hat{c}$. Subsequently, the final heatmap $\BO{h}_l$ is derived by the weighted aggregation of these feature maps, as shown in Eq. \ref{eq:cam_map}:

\begin{equation}
\label{eq:cam_map}
    \BO{h}_l = \operatorname{ReLU}\left( \sum_{k=1}^{C_l} \alpha_k^l \BO{f}_b^{l,k} \right).
\end{equation}

\textbf{Expert Knowledge Extraction.}
We select BioCLIP2 \citep{gu2025bioclip} as our domain expert among the many vision foundation models. Benefiting from contrastive pre-training on massive paired biological image-text data, it has a powerful classification prior.
During the entire training phase, we froze all parameters of the expert model to ensure that the robust classification prior of the expert is not confused by limited fine-tuned data.
As discussed before, complex background noise may lead to sub-optimal results for expert knowledge extraction.
To mitigate this, we design a masking strategy to filter out noise tokens based on the previously generated spatial cues.
Since the expert model follows the ViT architecture, it processes images by tokenizing them into a sequence of patches.
To ensure spatial alignment, we first interpolate all generated heatmaps $\BO{H}$ to match this resolution, denoted as $\tilde{\BO{h}}_l \in \mathbb{R}^{H_t \times W_t}$, where $H_t$ and $W_t$ are the resolution of the image processed by the expert model, for this expert, is $16 \times 16$.
Then, to avoid instability caused by continuous numerical variations, we binarize these heatmaps.
Specifically, for each layer, we retain only the pixels with the highest response values (i.e., top 50\%) as the informative foreground, while setting the rest to zero, formulated as:
\begin{equation}
\label{eq:mask_bin}
    \BO{m}_l^{(i,j)} =
    \begin{cases}
    1, & \text{if } \tilde{\BO{h}}_l^{(i,j)} \geq \operatorname{Median}(\tilde{\mathbf{h}}_l) \\
    0, & \text{otherwise}
    \end{cases}
\end{equation}
where $(i, j)$ represents the spatial coordinate indices within the token grid, and $\tilde{\mathbf{h}}_l^{(i,j)}$ denotes the response value of the interpolated heatmap at that specific location. We then apply these binary masks $\BO{M} = \{\BO{m}_1, \BO{m}_2, \BO{m}_3, \BO{m}_4\}$ to the input tokens of the expert model. By discarding tokens corresponding to the background, we extract four expert features $\{\BO{f}_e^1, \dots, \BO{f}_e^4\}$ that focus strictly on the discriminative tree parts derived from different backbone scales.
In addition to these masking features, we also extract the global feature $\BO{f}_e^0$ by feeding the original unmasked image to provide a robust fallback in case the CAM analysis fails.
Finally, we aggregate these features by a simple MLP to form the final expert decision $\BO{z}_e$, shown in Eq.\ref{eq:expert}.
\begin{equation}
\label{eq:expert}
    \BO{z}_e = \operatorname{MLP}\left( \operatorname{Concat}[\BO{f}_e^0, \BO{f}_e^1, \BO{f}_e^2, \BO{f}_e^3, \BO{f}_e^4] \right).
\end{equation}

\subsection{Uncertainty-guided Decision Calibration}

\begin{figure*}[t]
    \centering
    \includegraphics[width=1.0\linewidth]{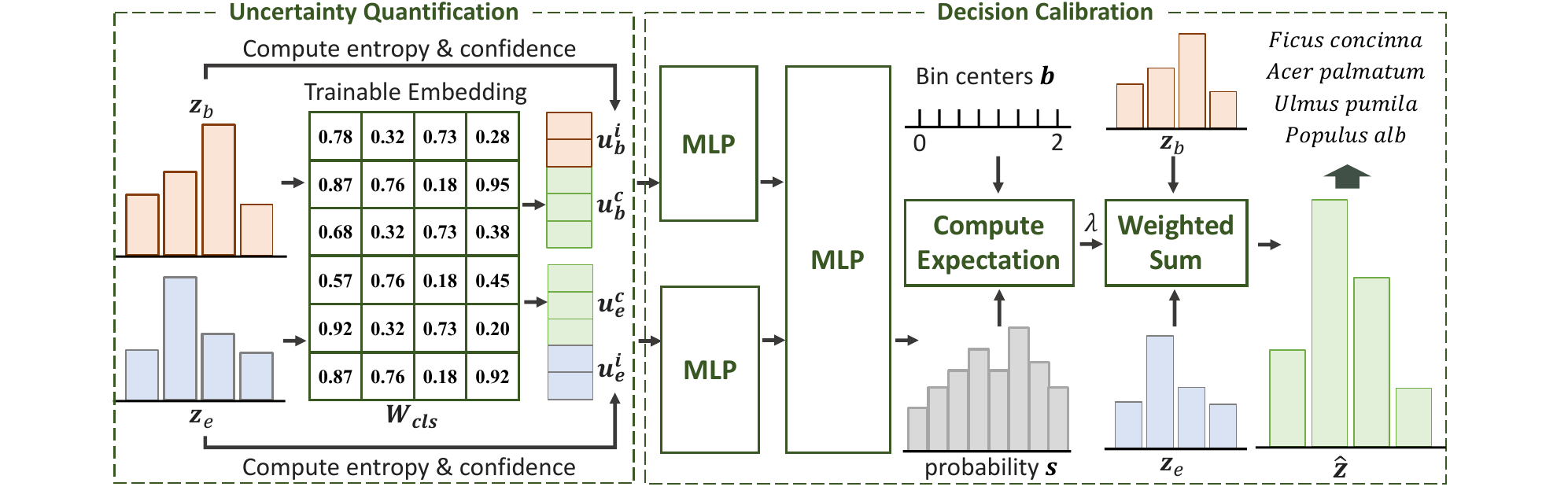}
    \caption{An illustration of the Uncertainty-guided Decision Calibration Module.}
    \label{fig:udcm}
\end{figure*}

Given the inherent bias of the local backbone trained on limited data, UDCM's primary objective is to leverage the domain expert's robust knowledge for rectification.
However, naively favoring a specific decision (e.g., tend to trust the expert) or treating both equally struggles to handle the complexities of real-world scenarios.
For example, when encountering a long-tail species rarely seen by the local backbone, the expert's robust prior becomes crucial, necessitating a high correction weight. Conversely, for a target heavily occluded by buildings, the frozen expert might be misled by background noise. In contrast, the local backbone, having adapted to such difficult scenes, actually proves more reliable. Driven by these observations, UDCM proposes to quantify the uncertainty of both agents to derive a dynamic calibration coefficient $\lambda$, as shown in Fig.\ref{fig:udcm}.

\textbf{Uncertainty Quantification.} We first decompose the uncertainty into two distinct dimensions: class-level uncertainty and instance-level uncertainty.
The former reflects the inherent difficulty of specific categories (e.g., long-tailed classes), whereas the latter captures the ambiguity of individual instances.
To model class-level uncertainty, we introduce a learnable class difficulty embedding $\BO{W}_{cls} \in \mathbb{R}^K$.
To incorporate the long-tail prior, we initialize $\BO{W}_{cls}$ using a smoothed inverse frequency strategy.
Specifically, let $N_c$ denote the number of training samples for category $c$. The initialization weight $w_c$ for the $c$-th class is computed as Eq.\ref{eq:initial}:
\begin{equation}
\label{eq:initial}
    \tilde{\mu}_c = 1 - \frac{N_c - N_{\min}}{N_{\max} - N_{\min}}, \quad w_c = (\tilde{\mu}_c)^\beta ,
\end{equation}
where $N_{\max}$ and $N_{\min}$ represent the maximum and minimum sample counts across all classes. This formulation maps the class frequencies to a normalized scale $[0, 1]$, and smooths the data distribution through hyperparameter $\beta$, ensuring that the model focuses more on scarce classes in the initial stage. These weights are then iteratively updated during training.
During the forward pass, we retrieve the weights corresponding to the Top-3 predicted categories for both the backbone $\BO{z}_b$ and the expert $\BO{z}_e$, respectively. These weights act as the class difficulty proxies, denoted as $\BO{u}_*^c \in \mathbb{R}^3$, formulated as Eq.\ref{eq:cls}:
\begin{equation}
\label{eq:cls}
    \BO{u}_*^c= \operatorname{Gather}(\BO{W}_{cls}, \operatorname{Top}_3(\BO{z}_{\ast})), \quad \text{where } \ast \in \{b, e\}.
\end{equation}
Here, $\operatorname{Top}_3(\cdot)$ identifies the indices of the three categories with the highest logits, and $\operatorname{Gather}(\cdot)$ retrieves the corresponding difficulty embeddings from $\BO{W}_{cls}$ based on these indices.
Simultaneously, we quantify instance-level uncertainty $\BO{u}_*^i \in \mathbb{R}^2$ using the probability distribution $\BO{p}_* = \operatorname{Softmax}(\BO{z}_*)$. As shown in Eq.\ref{eq:sample}, we compute the entropy and maximum confidence:
\begin{equation}
\label{eq:sample}
    \BO{u}_*^i = [ \underbrace{-\sum_{c=1}^{K} \BO{p}_{*,c} \log(\BO{p}_{*,c})}_{\text{Information Entropy}}, \;\; \underbrace{\max(\BO{p}_*)}_{\text{Prediction Confidence}} ],
\end{equation}
where $\BO{p}_{*,c}$ denotes the probability of the $c$-th class. Finally, we concatenate these proxies to obtain the holistic uncertainty embedding $\BO{u}_* = \operatorname{Concat}[\BO{u}_*^c, \BO{u}_*^i] \in \mathbb{R}^5$.

\textbf{Decision Calibration.}
Leveraging the quantified uncertainty $\BO{u}_b$ and $\BO{u}_e$, we generate a dynamic coefficient $\lambda$ to calibrate the final decision.
To ensure training stability, we bypass direct scalar regression and instead adopt a bin-based prediction strategy.
Specifically, we divide the potential weight range $[0, 2]$ uniformly into $N$ bins with centers $\BO{b} = \{b_1, \dots, b_N\}$.
We first project the uncertainty features via MLPs and then predict a probability distribution $\BO{s} \in \mathbb{R}^N$ over these bins. The scalar $\lambda$ is subsequently derived via expectation, formulated as Eq.\ref{eq:lambda}:
\begin{equation}
\label{eq:lambda}
\begin{aligned}
    & \BO{s} = \operatorname{Softmax}\left( \operatorname{MLP}( \operatorname{Concat}[\operatorname{MLP}_{proj}(\BO{u}_b), \operatorname{MLP}_{proj}(\BO{u}_e)] ) \right), \\
    & \lambda = \sum_{i=1}^{N} s_i \cdot b_i.
\end{aligned}
\end{equation}
Finally, we use this dynamic weight to rectify the backbone's decision. As shown in Eq.\ref{eq:calibration}, the expert's logits are scaled by $\lambda$ and added to the backbone, yielding the calibrated result $\hat{\BO{z}}$.
\begin{equation}
\label{eq:calibration}
    \hat{\BO{z}} = \BO{z}_b + \lambda \cdot \BO{z}_e.
\end{equation}

\subsection{Loss Function}
For this typical classification task, we utilize the standard Cross-Entropy (CE) loss, denoted as $\mathcal{L}_{ce}(\cdot)$, for optimization. Given the ground-truth label $y$, the total loss comprises three parts.
First, to ensure the classification capability of each branch, we compute the loss for the backbone logits $\BO{z}_b$ and the expert logits $\BO{z}_e$ individually:
\begin{equation}
    \mathcal{L}_b = \mathcal{L}_{ce}(\BO{z}_b, y), \quad \mathcal{L}_e = \mathcal{L}_{ce}(\BO{z}_e, y).
\end{equation}
For the final calibrated decision $\hat{\BO{z}}$, we introduce a dynamic re-weighting strategy. The total loss is formulated as Eq.\ref{eq:loss}:
\begin{equation}
\label{eq:loss}
    \mathcal{L}_{total} = \mathcal{L}_b + \mathcal{L}_e + \operatorname{sg}(\lambda) \cdot \mathcal{L}_{ce}(\hat{\BO{z}}, y),
\end{equation}
where $\operatorname{sg}(\cdot)$ represents the stop-gradient operation.
This design forces the model to focus on samples that require expert correction (i.e., $\lambda > 1$), while reducing the weight of samples where the expert is unreliable (i.e., $\lambda < 1$).
\section{Experiment}\label{sec:exp}
\begin{table*}[t]
\centering
\caption{Detailed statistics of the CU-Tree102 dataset. The 102 categories are sorted by sample frequency in descending order.}
\label{tab:dataset_stats}
\begin{threeparttable}
\begin{tabular}{lr | lr | lr}
\toprule
\textbf{Species Name} & \textbf{Count} & \textbf{Species Name} & \textbf{Count} & \textbf{Species Name} & \textbf{Count} \\
\midrule
\textit{Ficus microcarpa} & 286 & \textit{Albizia julibrissin} & 107 & \textit{Bauhinia purpurea} & 49 \\
\textit{Cocos nucifera} & 276 & \textit{Ligustrum lucidum} & 107 & \textit{Mangifera persiciforma} & 49 \\
\textit{Caryota maxima} & 237 & \textit{Metasequoia glyptostroboides} & 107 & \textit{Morus alba} & 49 \\
\textit{Trachycarpus fortunei} & 220 & \textit{Platycladus orientalis} & 105 & \textit{Dimocarpus longan} & 48 \\
\textit{Citrus maxima} & 219 & \textit{Yulania denudata} & 100 & \textit{Fraxinus chinensis} & 45 \\
\textit{Ginkgo biloba} & 199 & \textit{Picea meyeri} & 99 & \textit{Populus simonii} & 45 \\
\textit{Bauhinia blakeana} & 196 & \textit{Prunus cerasifera} & 99 & \textit{Morella rubra} & 44 \\
\textit{Archontophoenix alexandrae} & 186 & \textit{Cinnamomum japonicum} & 97 & \textit{Zelkova serrata} & 43 \\
\textit{Aesculus chinensis} & 180 & \textit{Sabina chinensis} & 95 & \textit{Picea koraiensis} & 41 \\
\textit{Salix babylonica} & 176 & \textit{Prunus salicina} & 94 & \textit{Populus canadensis} & 37 \\
\textit{Cedrus deodara} & 172 & \textit{Salix matsudana} & 93 & \textit{Elaeocarpus glabripetalus} & 35 \\
\textit{Euphorbia milii} & 172 & \textit{Broussonetia papyrifera} & 90 & \textit{Populus alba} & 35 \\
\textit{Triadica sebifera} & 160 & \textit{Firmiana simplex} & 89 & \textit{Populus davidiana} & 32 \\
\textit{Osmanthus fragrans} & 155 & \textit{Ficus altissima} & 88 & \textit{Michelia chapensis} & 31 \\
\textit{Livistona chinensis} & 151 & \textit{Platanus} & 88 & \textit{Pinus elliottii} & 31 \\
\textit{Cinnamomum camphora} & 149 & \textit{Celtis sinensis} & 87 & \textit{Quercus robur} & 30 \\
\textit{Magnolia grandiflora} & 142 & \textit{Liquidambar formosana} & 85 & \textit{Paulownia tomentosa} & 29 \\
\textit{Alstonia scholaris} & 139 & \textit{Picea asperata} & 85 & \textit{Toona sinensis} & 27 \\
\textit{Acer palmatum} & 137 & \textit{Sterculia lanceolata} & 76 & \textit{Celtis tetrandra} & 26 \\
\textit{Euonymus maackii} & 137 & \textit{Araucaria cunninghamii} & 72 & \textit{Grevillea robusta} & 26 \\
\textit{Catalpa ovata} & 135 & \textit{Bischofia polycarpa} & 71 & \textit{Ligustrum quihoui} & 26 \\
\textit{Photinia serratifolia} & 134 & \textit{Ficus virens} & 71 & \textit{Syringa villosa} & 26 \\
\textit{Lagerstroemia indica} & 131 & \textit{Eucommia ulmoides} & 69 & \textit{Pseudolarix amabilis} & 24 \\
\textit{Syringa reticulata} & 128 & \textit{Robinia pseudoacacia} & 69 & \textit{Pinus thunbergii} & 23 \\
\textit{Bombax ceiba} & 125 & \textit{Roystonea regia} & 69 & \textit{Ficus concinna} & 21 \\
\textit{Ceiba speciosa} & 123 & \textit{Melia azedarach} & 68 & \textit{Tilia amurensis} & 21 \\
\textit{Pterocarya stenoptera} & 119 & \textit{Erythrina variegata} & 67 & \textit{Prunus mandshurica} & 16 \\
\textit{Koelreuteria paniculata} & 118 & \textit{Pinus tabuliformis} & 67 & \textit{Populus tomentosa} & 15 \\
\textit{Pittosporum tobira} & 117 & \textit{Sapindus mukorossi} & 65 & \textit{Tilia mandshurica} & 15 \\
\textit{Ailanthus altissima} & 115 & \textit{Betula platyphylla} & 61 & \textit{Populus hopeiensis} & 14 \\
\textit{Ulmus pumila} & 115 & \textit{Elaeocarpus decipiens} & 55 & \textit{Phoebe zhennan} & 13 \\
\textit{Styphnolobium japonicum} & 112 & \textit{Rhododendron simsii} & 53 & \textit{Ulmus densa} & 13 \\
\textit{Ficus benjamina} & 111 & \textit{Dracontomelon duperreanum} & 51 & \textit{Populus cathayana} & 12 \\
\textit{Delonix regia} & 110 & \textit{Liriodendron chinense} & 51 & \textit{Populus nigra} & 11 \\
\bottomrule
\end{tabular}
\end{threeparttable}
\end{table*}

\subsection{Datasets}
\subsubsection{The CU-Tree102 Dataset}
To improve the diversity of existing benchmarks, we constructed the CU-Tree102 dataset with 102 tree species. We selected these species based on both scientific research and local surveys to cover the most common trees in China. Specifically, we referred to the study by \citep{wang2020urban}, which surveyed 35 major cities and identified 99 common street tree species. By combining these findings with reports from municipal forestry bureaus, we determined the final 102 categories as the targets for data collection. We used the Plant Photo Bank of China ( PPBC \footnote{\url{https://ppbc.iplant.cn/}}) as our primary source because every image was verified by experts, ensuring the accuracy of the labels.

However, the directly downloaded data did not meet the requirements for our task. A large number of images are of plant specimens from the laboratory, not actual tree appearances. To address this issue, we followed a strict cleaning process. First, we utilized a Vision-Language Model (VLM) to automatically filter out clearly irrelevant images. Next, we manually checked each image to ensure that it captured the entire tree structure.
Finally, we compiled a fine-grained classification data set that covers 9,134 in 102 classes, as shown in Tab.\ref{tab:dataset_stats}.

\subsubsection{Auxiliary Datasets}
To evaluate the robustness and generalization of our method, we used two additional datasets.

\textbf{RSTree.} The first dataset was derived from WHU-RSTree \citep{ding2025whu}. This source data provides paired point clouds and high-resolution panoramic images, featuring high-precision instance-level point cloud annotations and field-verified species information. To use this data for classification, we projected 3D masks onto the 2D images and cropped the individual trees based on their bounding boxes. We filtered out low-resolution samples, retaining only images with dimensions exceeding $224 \times 224$ pixels. The final dataset comprises 8,324 images across 23 tree species.

\textbf{Jekyll.} The second dataset is the Jekyll dataset \citep{yang2023urban}, which covers 23 categories with 4,804 images. We found that this dataset lacked diversity. Preliminary experiments showed that a simple ResNet-50 model could easily achieve 95.6\% accuracy on the official split, indicating a performance saturation that makes it unsuitable for distinguishing the effectiveness of fine-grained methods. Therefore, we only used it to test the generalization capability of our model.

\begin{table*}[t]
\centering
\caption{Main results on the CU-Tree102 dataset. Best results for each comparison pair are highlighted in \textbf{bold}, and $\uparrow$ indicates that higher is better. The last row summarizes the \textbf{average relative improvement} of our method across all baselines. \textbf{Acc}: Top-1 Accuracy. \textbf{mP/mR/mF1}: Macro metrics. \textbf{wP/wR/wF1}: Weighted metrics.}
\label{tab:main_results}
\begin{tabular}{l|c|c|ccc|ccc}
\toprule
\multirow{2}{*}{\textbf{Method}} & \textbf{Params} & \textbf{Acc $\uparrow$} & \multicolumn{3}{c|}{\textbf{Macro Avg. (\%) $\uparrow$}} & \multicolumn{3}{c}{\textbf{Weighted Avg. (\%) $\uparrow$}} \\
 & \textbf{(M)} & \textbf{(\%) } & \textbf{mP} & \textbf{mR} & \textbf{mF1} & \textbf{wP} & \textbf{wR} & \textbf{wF1} \\
\midrule
Expert-FT & 303.77 & 81.61 & 74.43 & 70.20 & 70.93 & 81.28 & 81.61 & 80.87 \\
\midrule
\multicolumn{9}{l}{\textit{\textbf{General Deep Learning Backbones}}} \\
\midrule
ResNet-50 \citep{he2016deep} & 23.72 & 68.26 & 55.28 & 53.43 & 53.33 & 66.99 & 68.26 & 66.98 \\
\textbf{+ Ours} & +0.08 & \textbf{77.96} & \textbf{69.81} & \textbf{66.73} & \textbf{67.25} & \textbf{77.88} & \textbf{77.96} & \textbf{77.42} \\
\cmidrule{1-9}
ViT-Base \citep{dosovitskiy2020image} & 86.14 & 73.30 & 64.11 & 60.67 & 61.15 & 73.11 & 73.30 & 72.65 \\
\textbf{+ Ours} & +0.08 & \textbf{78.07} & \textbf{69.83} & \textbf{65.81} & \textbf{66.60} & \textbf{77.40} & \textbf{78.07} & \textbf{77.20} \\
\cmidrule{1-9}
Swin-Base \citep{liu2021swin} & 109.37 & 79.33 & 69.62 & 67.72 & 67.78 & 78.76 & 79.33 & 78.64 \\
\textbf{+ Ours} & +0.08 & \textbf{83.44} & \textbf{75.83} & \textbf{72.86} & \textbf{73.36} & \textbf{82.78} & \textbf{83.44} & \textbf{82.68} \\
\midrule
\multicolumn{9}{l}{\textit{\textbf{State-of-the-Art FGVC Methods}}} \\
\midrule
HERBS \citep{chou2023fine} & 161.65 & 82.98 & 74.05 & 70.91 & 71.26 & 81.98 & 82.98 & 81.90 \\
\textbf{+ Ours} & +0.08 & \textbf{85.61} & \textbf{78.29} & \textbf{76.11} & \textbf{76.44} & \textbf{85.22} & \textbf{85.61} & \textbf{85.06} \\
\cmidrule{1-9}
MPSA \citep{wang2024multi} & 97.26 & 80.84 & 65.07 & 65.50 & 63.80 & \textbf{87.21} & 80.84 & 83.32 \\
\textbf{+ Ours} & +0.08 & \textbf{83.31} & \textbf{70.44} & \textbf{74.45} & \textbf{70.41} & 87.00 & \textbf{83.31} & \textbf{84.67} \\
\cmidrule{1-9}
CGL \citep{cgl} & 128.19 & 81.47 & 72.14 & 70.37 & 70.34 & 81.32 & 81.47 & 81.03 \\
\textbf{+ Ours} & +0.08 & \textbf{86.65} & \textbf{79.83} & \textbf{77.35} & \textbf{77.87} & \textbf{86.31} & \textbf{86.65} & \textbf{86.17} \\
\midrule
\textbf{Average Improvement.} & / & \textbf{+6.42\%} & \textbf{+11.46\%} & \textbf{+11.98\%} & \textbf{+11.93\%} & \textbf{+6.18\%} & \textbf{+6.42\%} & \textbf{+6.47\%} \\
\bottomrule
\end{tabular}%
\end{table*}

\subsection{Evaluation Setting}
\subsubsection{Implementation Details}
We divided datasets into training, validation, and testing sets with a ratio of 8:1:1. Notably, to ensure reliable evaluation for long-tailed classes, we adjusted this split to guarantee that each category in the test set contained at least five samples. Following previous works \citep{wang2024multi,cgl}, we resized the images to $510 \times 510$ pixels and cropped them to $384 \times 384$ pixels. During training, we applied data augmentation techniques, including random flipping, blurring, and sharpness adjustments. We used the SGD optimizer with an initial learning rate of 0.0005 and trained the model for 100 epochs on a single NVIDIA RTX 4090 GPU. For the hyperparameters, we set $\beta=2.0$ and the number of bins $N=8$. To ensure reproducibility, our code, models, and datasets are publicly available at \url{https://github.com/WHU-USI3DV/TreeCLS}.

\subsubsection{Evaluation Metrics}
To comprehensively assess the performance, we employed four standard metrics: Top-1 Accuracy (Acc), Precision, Recall, and F1-score (F1).
Due to the imbalanced distribution, we reported both \textbf{Macro-average} and \textbf{Weighted-average} metrics. The Macro-average mode treats all classes equally to reflect the performance on tail categories, while the Weighted-average mode weights the score based on the number of samples in each category to reflect the overall system performance. For brevity, we denote the Macro-average and Weighted-average metrics with the prefixes m and w, respectively (e.g., mF1 and wF1).

\begin{figure*}[t]
    \centering
    \includegraphics[width=1.0\linewidth]{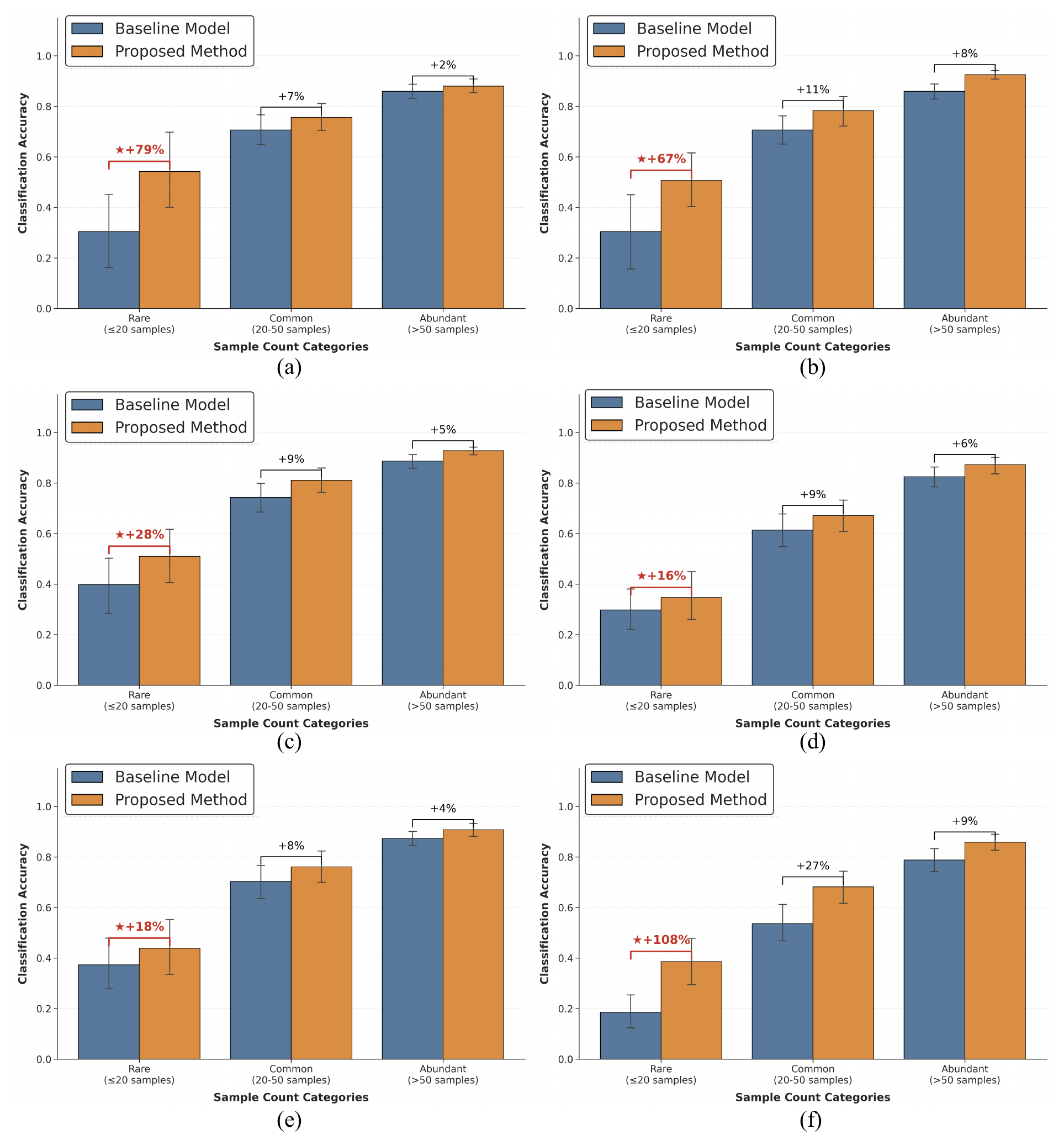}
    \caption{Accuracy comparison grouped by sample counts. We evaluate the proposed EKDC-Net against three state-of-the-art fine-grained methods: (a) MPSA (b) HERBS, and (c) CGL; and three standard backbones: (d) ViT-Base, (e) Swin-Base, and (f) ResNet-50. The proposed method (orange) consistently outperforms the baseline (blue). Percentages represent relative improvement, and * indicates the intervals with the greatest increase.}
    \label{fig:longtail}
\end{figure*}

\begin{figure*}[t]
    \centering
    \includegraphics[width=1.0\linewidth]{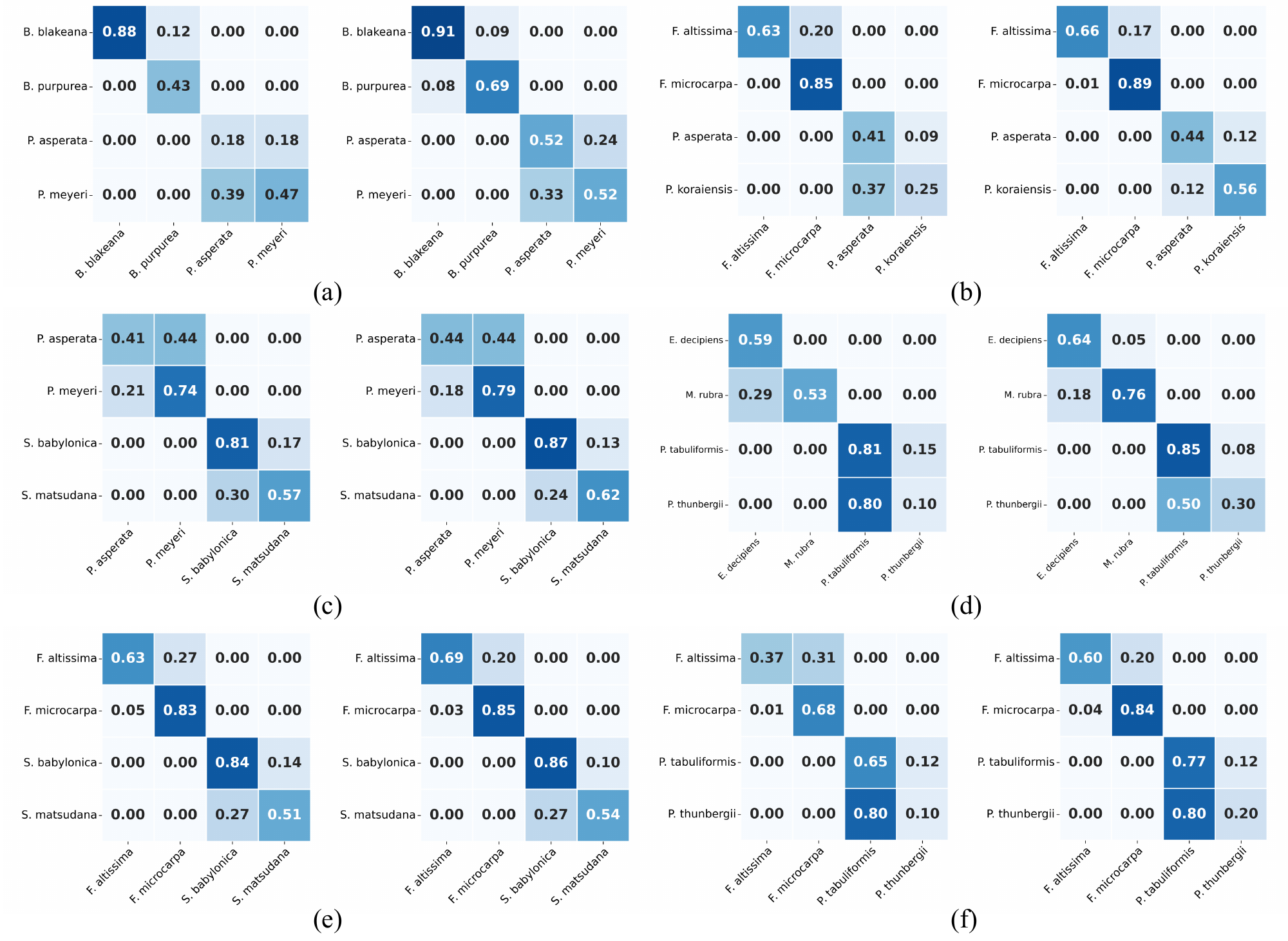} 
    \caption{Visualization of confusion matrices for the most challenging categories. Each subplot displays the performance of the baseline (left) compared to our EKDC-Net (right): (a) MPSA, (b) HERBS, (c) CGL, (d) ViT-Base, (e) Swin-Base, and (f) ResNet-50. The Darker colors indicate larger values. }
    \label{fig:confuse}
\end{figure*}

\subsubsection{Comparison Methods}

To validate the effectiveness of our proposed framework, we conducted experiments against three distinct categories of baselines.
\textbf{General Backbones.}
First, we selected some widely used representative deep learning backbones. These included ResNet-50 \citep{he2016deep} for CNN architectures, ViT-Base \citep{dosovitskiy2020image} and Swin-Base \citep{liu2021swin} for Transformer architectures.
\textbf{FGVC Models.}
The lack of publicly available code for existing forestry-specific methods \citep{dong2025multiscale} hinders fair comparison. As an alternative, we selected general FGVC methods, such as HERBS \citep{chou2023fine}, MPSA \citep{wang2024multi}, and CGL \citep{cgl}, to evaluate existing capabilities in identifying fine-grained tree species.
\textbf{Expert Fine-tuning.}
Finally, to verify the value of our expert guidance fusion strategy, we implemented full parameter fine-tuning on the expert model (denoted as Expert-FT). This served as a strong baseline to determine whether our collaborative framework offers better performance than simple retraining.

\subsection{Experimental Performance}
\subsubsection{Results on CU-Tree102}
To ensure a fair comparison, we retrained all comparison methods on the CU-Tree102 dataset using their official open-source code and configurations.
The comprehensive quantitative results presented in Tab.\ref{tab:main_results} demonstrate that our proposed EKDC-Net effectively boosts the performance of both traditional deep learning backbones and advanced FGVC architectures designed for discriminative feature learning.
Specifically, our framework elevates the Top-1 Accuracy of existing backbones by an average of 6.42\% and the overall precision by 11.46\%.

Notably, we observed that the improvement of macro indicators is significantly greater than that of weighted indicators.
Taking ResNet-50 as an instance, while the weighted F1 score increases by 15.59\%, the macro F1 score surges by 26.10\%.
Compared to the weighted indicators, the macro indicators treat all categories equally and are not affected by the number of samples in each category. This additional significant improvement proves that our method effectively corrects the model's bias towards the head categories, thereby providing a crucial discriminative basis for the challenging long-tail categories. To further validate this capability on tail classes, we divided the dataset into three subsets based on sample frequency: Rare ($<$20 samples), Common (20-50 samples), and Abundant ($>$50 samples).
We then calculated the accuracy improvement range for each segment.
The results in Fig.\ref{fig:longtail} indicate that the performance improvement of our framework is particularly outstanding in the few-shot regime, confirming its excellent performance for long-tail classes.

Beyond mitigating data imbalance, we further validated the model's capability to distinguish confusing categories that pose significant challenges to backbone networks.
Specifically, we identified the Top-4 most confusing species pairs for each baseline model and compared the normalized confusion matrices between the baselines and our framework on these specific classes, as illustrated in Fig.\ref{fig:confuse}.
As observed, the baseline models exhibit severe ambiguity when distinguishing visually similar subspecies.
In contrast, our EKDC-Net effectively suppresses these mutual misclassifications, resulting in a clearer diagonal structure in the confusion matrix.
This improvement suggests that our method not only mitigates the bias from long-tailed distributions but also effectively refines the decision boundaries for hard-to-distinguish fine-grained classes.

Furthermore, we evaluated the efficacy of our well-designed modules by comparing it against the fully fine-tuned expert model.
The results in Table \ref{tab:main_results} show that our collaborative strategy outperforms the full-parameter training of expert models.
This confirms that the improvement in judgment ability arises from the efficient design of expert knowledge extraction and utilization, rather than simply updating the parameters of the experts.

Finally, we assessed the additional overhead introduced by our framework.
As indicated in the "Params" column, EKDC-Net introduces only \textbf{0.08M} additional learnable parameters.
These parameters primarily comprise the lightweight linear projection layers in LPKEM and the uncertainty estimation MLPs in UDCM.
Compared to parameter-heavy backbones (e.g., HERBS with 161.65M parameters), this increase is negligible (i.e., less than 0.1\%).
This parameter efficiency confirms that the significant performance improvements result from the integration of high-quality expert knowledge, rather than simply increasing model capacity.

\begin{table*}[t]
\centering
\caption{Comparison Results on the auxiliary \textbf{RSTree} dataset. All models were retrained.}
\label{tab:rstree_results}
\begin{tabular}{l|c|ccc|ccc}
\toprule
\multirow{2}{*}{\textbf{Method}} & \textbf{Acc $\uparrow$} & \multicolumn{3}{c|}{\textbf{Macro Avg. (\%) $\uparrow$}} & \multicolumn{3}{c}{\textbf{Weighted Avg. (\%) $\uparrow$}} \\
 & \textbf{(\%)} & \textbf{mP} & \textbf{mR} & \textbf{mF1} & \textbf{wP} & \textbf{wR} & \textbf{wF1} \\
\midrule
\multicolumn{8}{l}{\textit{\textbf{Standard Backbones}}} \\
\midrule
ResNet-50 & 88.01 & 68.59 & 64.84 & 62.20 & 88.26 & 88.01 & 87.62 \\
\textbf{+ Ours} & \textbf{90.19} & \textbf{79.02} & \textbf{73.67} & \textbf{73.63} & \textbf{90.87} & \textbf{90.19} & \textbf{90.10} \\
\cmidrule{1-8}
ViT-Base & 89.46 & 83.04 & 65.43 & 68.73 & 90.09 & 89.46 & 89.11 \\
\textbf{+ Ours} & \textbf{91.01} & \textbf{87.33} & \textbf{71.58} & \textbf{74.36} & \textbf{91.60} & \textbf{91.01} & \textbf{90.66} \\
\cmidrule{1-8}
Swin-Base & 92.01 & 83.98 & 73.33 & 75.76 & 92.25 & 92.01 & 91.77 \\
\textbf{+ Ours} & \textbf{92.92} & \textbf{88.98} & \textbf{73.88} & \textbf{77.97} & \textbf{93.19} & \textbf{92.92} & \textbf{92.66} \\
\midrule
\multicolumn{8}{l}{\textit{\textbf{FGVC Method}}} \\
\midrule
CGL & 65.67 & 5.71 & 7.92 & 6.61 & 48.12 & 65.67 & 55.32 \\
\textbf{+ Ours} & \textbf{87.83} & \textbf{55.49} & \textbf{49.68} & \textbf{49.63} & \textbf{86.51} & \textbf{87.83} & \textbf{86.74} \\
\midrule
\textbf{Average Improvement.} & \textbf{+9.74\%} & \textbf{+224.6\%} & \textbf{+137.8\%} & \textbf{+170.1\%} & \textbf{+21.36\%} & \textbf{+9.74\%} & \textbf{+15.59\%} \\
\bottomrule
\end{tabular}%
\end{table*}

\subsubsection{Results on RSTree}

To evaluate the robustness of our framework under extreme data imbalance, we extended our experiments to the RSTree dataset.
Unlike CU-Tree102, RSTree originated directly from authentic field measurement labels rather than being specifically constructed for classification benchmarks.
This allows RsTree to capture the distribution of tree species in real-world forestry scenarios, with abundant sample sizes for dominant species and extremely few for rare species.
Specifically, RSTree exhibits a much more severe long-tailed distribution compared to curated datasets: the most frequent categories contain nearly \textbf{2,000} samples, whereas the rarest species have as few as \textbf{5} samples.
We retrained all baseline models (including standard backbones and the SOTA FGVC method CGL) and our EKDC-Net from scratch on this challenging dataset.
The quantitative results are summarized in Table \ref{tab:rstree_results}.

Under this extreme imbalance, the CGL baseline experienced a catastrophic collapse, achieving a Macro F1-score of only \textbf{6.61\%}.
This indicates that without proper guidance, the model is completely overwhelmed by head classes and fails to identify nearly all minority species.
In contrast, our EKDC-Net, by calibrating the biased decisions of the backbone network, successfully boosted the model's Top-1 accuracy from 65.67\% to 87.83\%, and the macro F1 score recovered to 49.63\%.

For standard deep learning backbones (ResNet, ViT, Swin), our method consistently provides robust improvements. Specifically, EKDC-Net improves the Top-1 Accuracy of ResNet-50, ViT-Base, and Swin-Base by \textbf{2.47\%}, \textbf{1.73\%}, and \textbf{0.99\%}, respectively.
Despite the high saturation of head class accuracy, our approach still achieves further gains, demonstrating strong generalization capabilities across different architectures.

Consistent with previous findings, the improvement of macro indicators is significantly greater than that of weighted indicators. On average, even excluding the outlier CGL, the backbone still shows a substantial average gain of 9.8\%. This huge difference confirms that our method effectively focuses on tail samples, significantly improving the overall identification capability of the system.

\begin{table*}[t]
\centering
\caption{Generalization results on the Jekyll dataset. Models pre-trained on CU-Tree102 were directly applied to Jekyll without fine-tuning.}
\label{tab:jekyll_generalization}
\begin{tabular}{l|c|ccc|ccc}
\toprule
\multirow{2}{*}{\textbf{Method}} & \textbf{Acc $\uparrow$} & \multicolumn{3}{c|}{\textbf{Macro Avg. (\%) $\uparrow$}} & \multicolumn{3}{c}{\textbf{Weighted Avg. (\%) $\uparrow$}} \\
 & \textbf{(\%)} & \textbf{mP} & \textbf{mR} & \textbf{mF1} & \textbf{wP} & \textbf{wR} & \textbf{wF1} \\
\midrule
\multicolumn{8}{l}{\textit{\textbf{Standard Backbones}}} \\
\midrule
ResNet-50 & 26.83 & 58.84 & 27.17 & 31.03 & 58.48 & 26.83 & 29.99 \\
\textbf{+ Ours} & \textbf{39.22} & \textbf{66.81} & \textbf{39.08} & \textbf{42.83} & \textbf{63.65} & \textbf{39.22} & \textbf{41.56} \\
\cmidrule{1-8}
ViT-Base & 26.38 & 46.41 & 25.80 & 28.88 & 44.63 & 26.38 & 28.43 \\
\textbf{+ Ours} & \textbf{35.32} & \textbf{62.79} & \textbf{35.27} & \textbf{38.02} & \textbf{62.37} & \textbf{35.32} & \textbf{36.74} \\
\cmidrule{1-8}
Swin-Base & 30.05 & 56.09 & 28.39 & 30.80 & 55.52 & 30.05 & 31.40 \\
\textbf{+ Ours} & \textbf{44.27} & \textbf{69.79} & \textbf{44.43} & \textbf{49.28} & \textbf{68.37} & \textbf{44.27} & \textbf{48.40} \\
\midrule
\multicolumn{8}{l}{\textit{\textbf{State-of-the-Art FGVC Methods}}} \\
\midrule
HERBS & 52.98 & 80.56 & 52.79 & 60.04 & 80.14 & 52.98 & 59.75 \\
\textbf{+ Ours} & \textbf{59.63} & \textbf{82.14} & \textbf{60.11} & \textbf{67.01} & \textbf{81.69} & \textbf{59.63} & \textbf{66.48} \\
\cmidrule{1-8}
CGL & 51.83 & 81.08 & 50.22 & 58.04 & 79.84 & 51.83 & 59.13 \\
\textbf{+ Ours} & \textbf{58.49} & \textbf{83.34} & \textbf{60.09} & \textbf{65.07} & \textbf{81.54} & \textbf{58.49} & \textbf{62.66} \\
\midrule
\textbf{Average Improvement. } & \textbf{+30.56\%} & \textbf{+15.60\%} & \textbf{+34.11\%} & \textbf{+30.68\%} & \textbf{+15.16\%} & \textbf{+30.56\%} & \textbf{+27.84\%} \\
\bottomrule
\end{tabular}%
\end{table*}

\subsection{Generalization Test}

To evaluate the generalization ability of our framework, we performed an evaluation on the Jekyll dataset.
Unlike the previous experiments involving retraining, here we directly applied the models pre-trained on CU-Tree102 to the Jekyll test set without any fine-tuning.
Since only 21 out of the 23 categories in Jekyll overlap with CU-Tree102, we only evaluated the performance of these 21 shared species, and any sample predicted to be in non-overlapping categories was strictly considered as misclassified.

As detailed quantitative results in Tab.\ref {tab:jekyll_generalization} show, significant domain gaps cause the standard baseline to suffer severe performance degradation.
For instance, the accuracies of ResNet-50 and ViT-Base dropped to approximately 26\%, indicating that they likely overfitted the specific data distribution of CU-Tree102.
In contrast, our EKDC-Net demonstrated remarkable robustness.
By integrating expert knowledge, we improved the Top-1 Accuracy of ResNet-50 and Swin-Base by 46.18\% and 47.32\% relative to their baselines, respectively.
This result provides compelling evidence for our core motivation: our design is not simply to force the model to overfit the local distribution of the training set, but rather to use the foundation model as an external expert to guide the backbone to focus on intrinsic, transferable classification features (such as leaf morphology and branching patterns). This capability allows the model to maintain high discriminability even when deployed in unseen environments.

\begin{table*}[t]
\centering
\caption{Component contribution analysis on the CU-Tree102 dataset. \textbf{FM}: Foundation Model (Expert).}
\label{tab:ablation_component}
\begin{tabular}{ccc|c|ccc|ccc}
\toprule
\multicolumn{3}{c|}{\textbf{Components}} & \textbf{Acc $\uparrow$} & \multicolumn{3}{c|}{\textbf{Macro Avg. (\%) $\uparrow$}} & \multicolumn{3}{c}{\textbf{Weighted Avg. (\%) $\uparrow$}} \\
\textbf{FM} & \textbf{UDCM} & \textbf{LPKEM} & \textbf{(\%)} & \textbf{mP} & \textbf{mR} & \textbf{mF1} & \textbf{wP} & \textbf{wR} & \textbf{wF1} \\
\midrule
 &  &  & 81.47 & 72.14 & 70.37 & 70.34 & 81.32 & 81.47 & 81.03 \\
\checkmark &  &  & 85.75 & 73.50 & 72.87 & 72.61 & 83.64 & 85.75 & 84.30 \\
\checkmark & \checkmark &  & 85.80 & 78.15 & 74.57 & 74.67 & 85.17 & 85.80 & 84.81 \\
\checkmark & \checkmark & \checkmark & \textbf{86.65} & \textbf{79.83} & \textbf{77.35} & \textbf{77.87} & \textbf{86.31} & \textbf{86.65} & \textbf{86.17} \\
\bottomrule
\end{tabular}
\end{table*}
\begin{table*}[t]
\centering
\caption{Comparison with different fusion strategies on CU-Tree102. }
\label{tab:fusion}
\begin{tabular}{l|c|ccc|ccc}
\toprule
\multirow{2}{*}{\textbf{Fusion Strategy}} & \textbf{Acc $\uparrow$} & \multicolumn{3}{c|}{\textbf{Macro Avg. (\%) $\uparrow$}} & \multicolumn{3}{c}{\textbf{Weighted Avg. (\%) $\uparrow$}} \\
 & \textbf{(\%)} & \textbf{mP} & \textbf{mR} & \textbf{mF1} & \textbf{wP} & \textbf{wR} & \textbf{wF1} \\
\midrule
\textit{Logit-level Fusion} & & & & & & & \\
MLP Fusion & 79.99 & 70.59 & 68.95 & 69.05 & 79.82 & 79.99 & 79.56 \\
Attention Fusion & 78.51 & 69.66 & 67.10 & 67.49 & 79.87 & 78.51 & 78.75 \\
\midrule
\textit{Feature-level Fusion} & & & & & & & \\
MLP Fusion & 81.20 & 72.27 & 70.08 & 70.46 & 80.93 & 81.20 & 80.77 \\
Attention Fusion & 80.54 & 71.14 & 69.11 & 69.34 & 80.18 & 80.54 & 79.99 \\
\midrule
\textbf{Ours (UDCM)} & \textbf{86.65} & \textbf{79.83} & \textbf{77.35} & \textbf{77.87} & \textbf{86.31} & \textbf{86.65} & \textbf{86.17} \\
\bottomrule
\end{tabular}
\end{table*}

\begin{figure*}[t]
    \centering
    \includegraphics[width=1.0\linewidth]{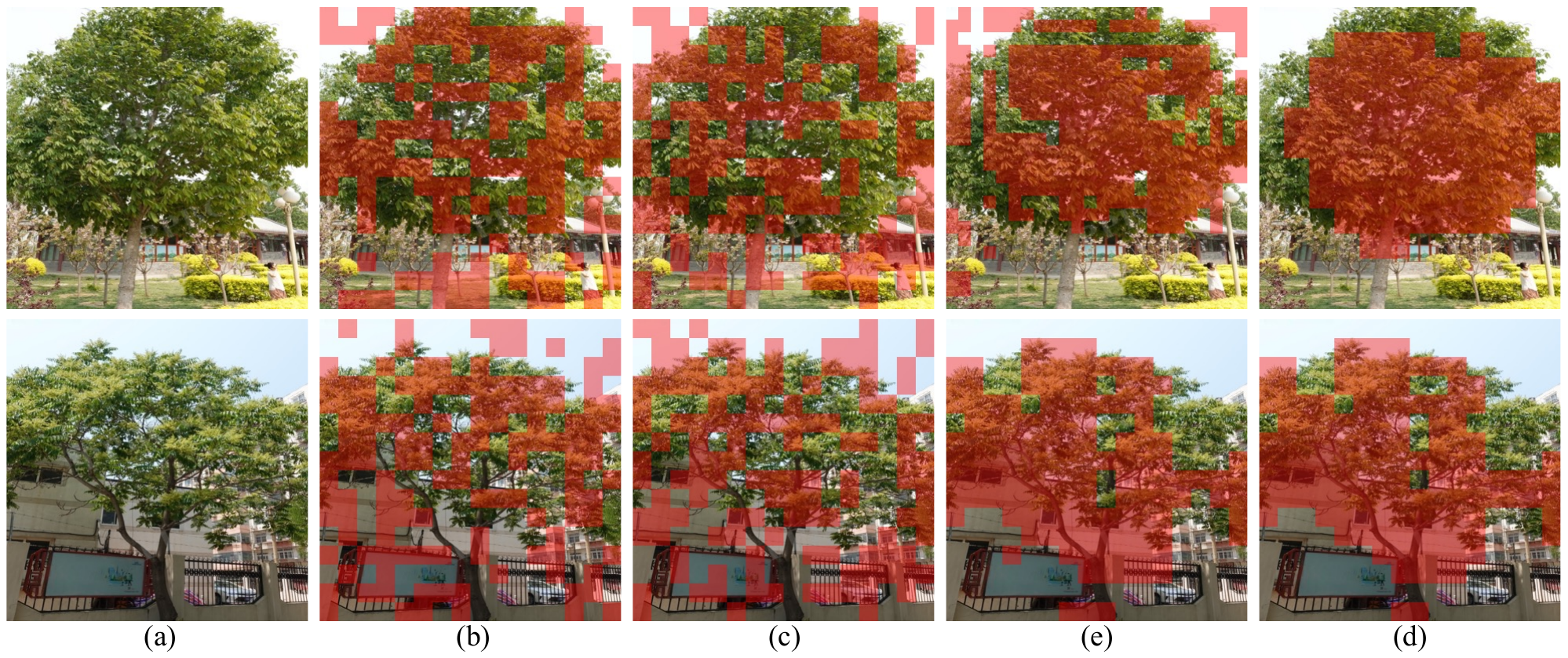} 
    \caption{Visualization of the hierarchical binary masks generated by LPKEM. (a) Input image. (b)-(e) Binary masks ($\BO{m}_1$ to $\BO{m}_4$) derived from the multi-scale feature maps $\BO{F}_b$. The red overlays indicate the regions preserved for the expert model.}
    \label{fig:lpkem_vis}
\end{figure*}

\section{Discussion}
\subsection{Component Contribution Analysis}
To evaluate the contribution of each component within our framework, we conducted a progressive ablation study on the CU-Tree102 dataset using the CGL backbone.
We established three variants by systematically excluding components:
1) Removing the masking strategy in LPKEM and feeding the full image to the expert.
2) Replacing the UDCM module with a simple average fusion.
3) Discarding the expert model entirely (i.e., reverting to the original CGL baseline).

The quantitative results detailed in Table \ref{tab:ablation_component} show that each proposed module plays a critical role in improving the model's performance.
First, the introduction of expert knowledge yielded the most significant improvement, increasing the Top-1 accuracy from 81.47\% to 85.75\%. This confirms that robust general knowledge effectively corrects decisions biased by local data.
Furthermore, we observe a critical phenomenon when removing UDCM and LPKEM: the decline in Macro metrics (e.g., mP and mR) significantly exceeds that of Top-1 Accuracy.
Specifically, without the UDCM and LPKEM strategies, Macro Precision drops by \textbf{6.33\%}, whereas Accuracy decreases by only \textbf{0.90\%}.
This significant discrepancy demonstrates that simply incorporating expert knowledge is insufficient to address the long-tailed challenges inherent in local datasets.
In contrast, LPKEM refines knowledge extraction by generating precise masks to filter out background noise. As visually verified in Fig.\ref{fig:lpkem_vis}, for images with complex urban backgrounds (e.g., buildings and fences), the generated multi-scale masks ($\BO{m}_1$ to $\BO{m}_4$) effectively isolate the tree structure.
Simultaneously, UDCM maintains performance on hard-to-distinguish fine-grained classes by allowing the model to adaptively rely on expert priors when local predictions exhibit high uncertainty.
The effective integration of these two modules enables the model to achieve the best performance across all classification metrics.

\begin{figure*}[t]
    \centering
    \includegraphics[width=0.6\linewidth]{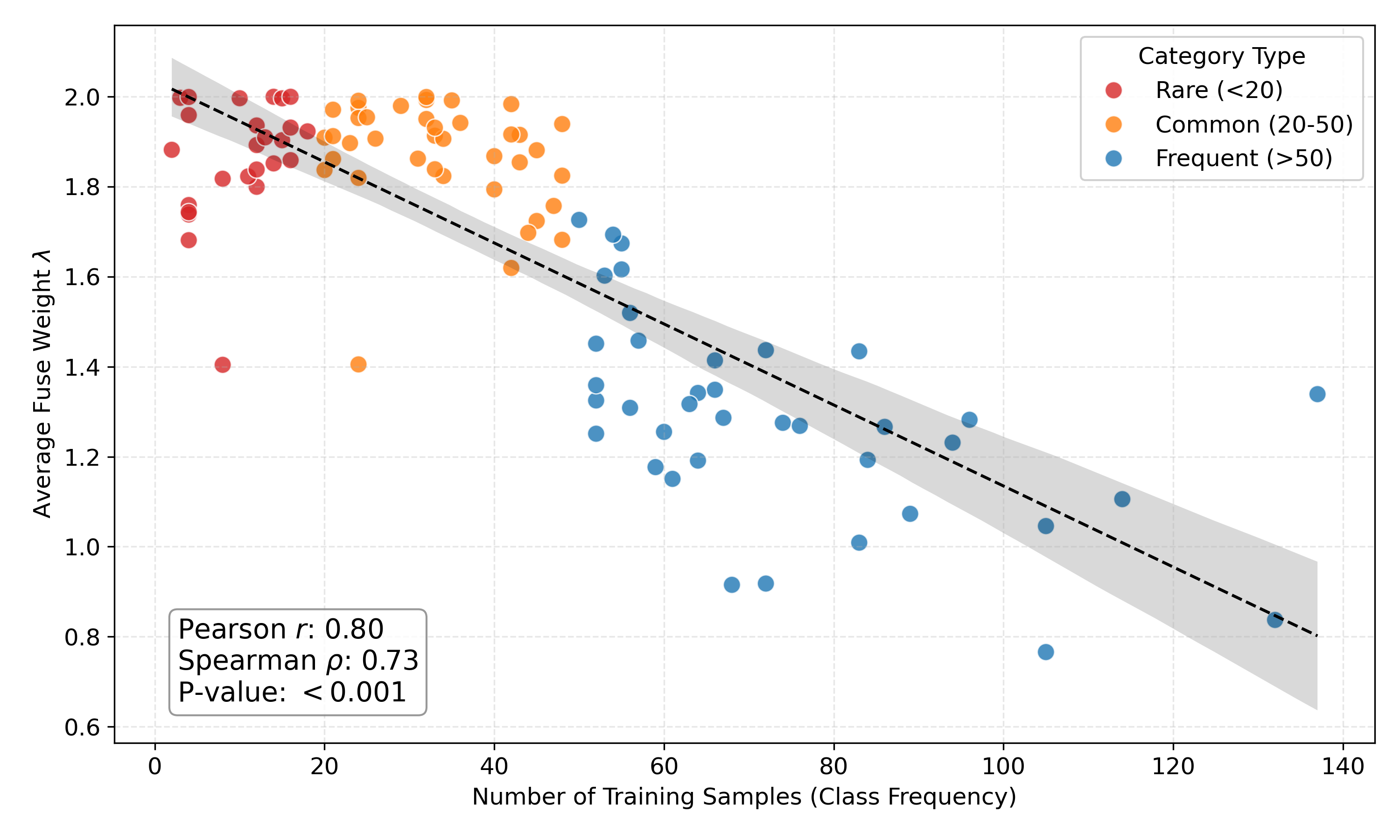}
    \caption{Correlation analysis between class frequency and learned calibration weight $\lambda$.}
    \label{fig:long_tail_corr}
\end{figure*}

\begin{figure*}[t]
    \centering
    \includegraphics[width=1.0\linewidth]{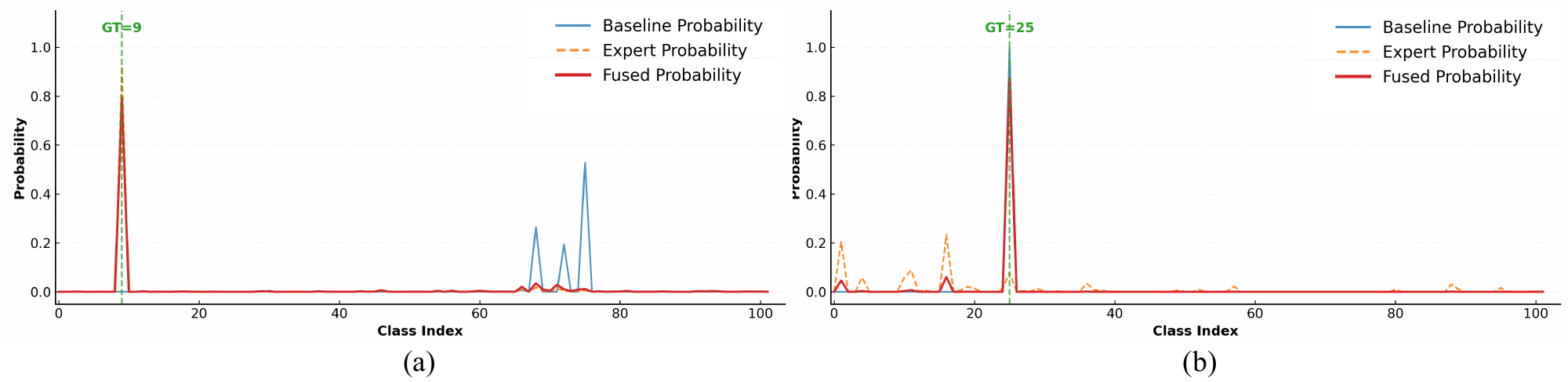}
    \caption{Visualization of the adaptive decision calibration process by UDCM. (a) Expert Correction ($\lambda=1.98$): Due to the high uncertainty of the baseline model, the UDCM assigned a fusion weight of 1.98 to correct the classification decision. (b) Noise Suppression ($\lambda=0.12$): When experts make incorrect predictions, UDCM generates a low weight of 0.12 to suppress the noise from experts while retaining the correct result. }
    \label{fig:case_study}
\end{figure*}

\begin{figure*}[t]
    \centering
    \includegraphics[width=0.6\linewidth]{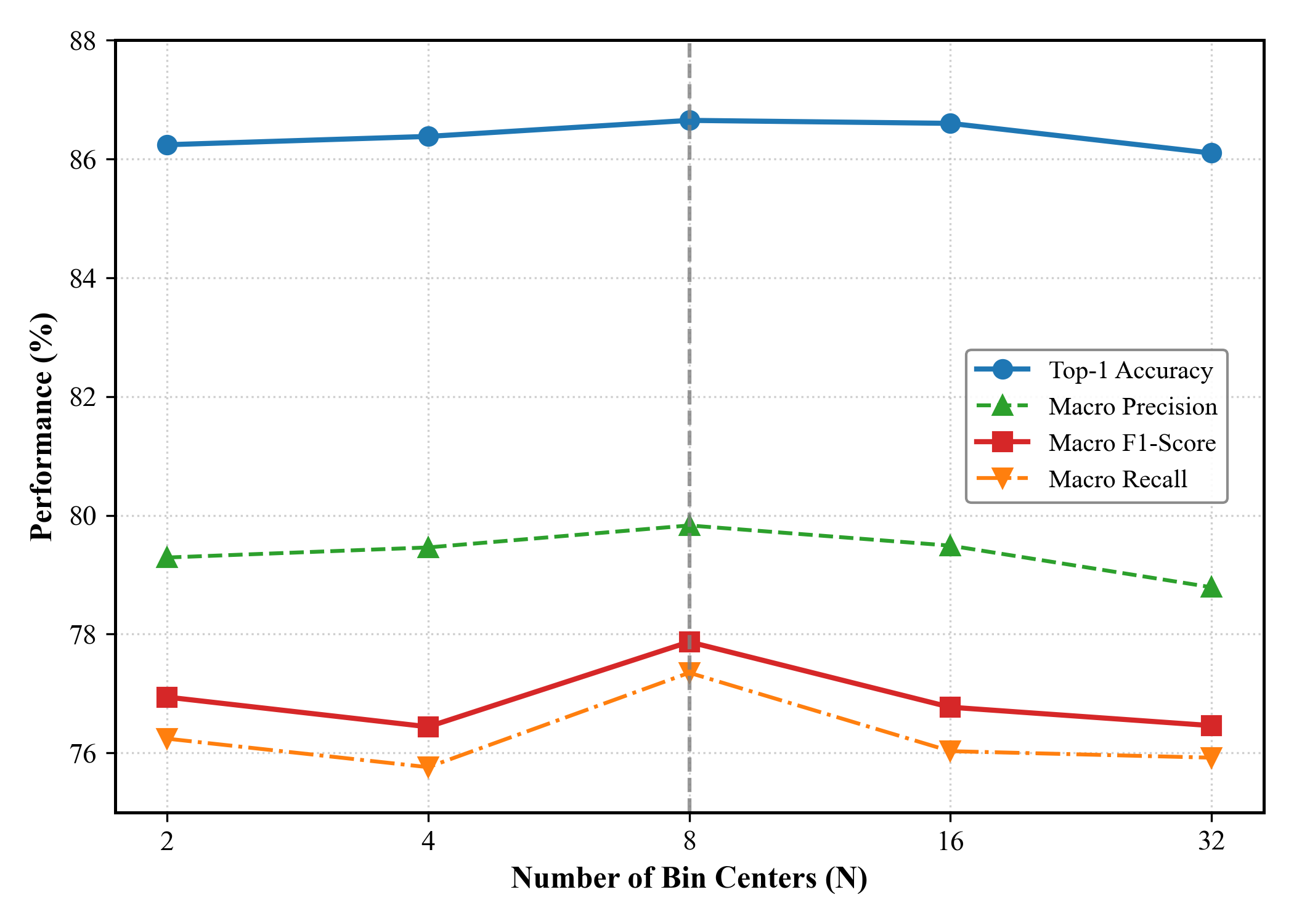}
    \caption{Parameter sensitivity analysis of $N$ on the CU-Tree102 dataset. The vertical dashed line marks the optimal setting.}
    \label{fig:k_sensitivity}
\end{figure*}

\subsection{Analysis of Decision Calibration Strategy}

To demonstrate the superiority of our Uncertainty-guided Decision Calibration Module (UDCM), we compare it against four representative fusion paradigms:
\textbf{Logit-level fusion.} It uses MLP or Attention Block to fuse expert predictions $\BO{z}_e$ and local predictions $\BO{z}_b$, respectively.
\textbf{Feature-level fusion.} Expert features are concatenated with local features and processed by MLP or an attention layer.

The quantitative results in Tab.\ref{tab:fusion} show that standard fusion methods perform significantly worse than our UDCM.
Specifically, feature-level fusion strategies achieve an accuracy of only $\sim$81\%.
This limitation mainly stems from the mismatch in feature representations.
The expert features reside in a general semantic space, whereas the local features are optimized for the specific task distribution.
Forcing them to fuse without fine-tuning disrupts the integrity of both representations.
Furthermore, logit-level fusion strategies also fail to yield satisfactory results (Acc $<$ 80\%).
The core reason is the lack of effective guidance.
Without modeling uncertainty, standard MLPs or attention mechanisms struggle to dynamically determine which model to trust, leading to suboptimal compromises rather than effective calibration.
In contrast, our UDCM achieves superior classification performance by explicitly quantifying both class-level and instance-level uncertainty.

To understand the internal mechanism of UDCM, we analyzed the statistical distribution of the learned weight $\lambda$.
Specifically, we analyzed the correlation between class frequency (i.e., training sample count) and $\lambda$.
As illustrated in Fig.\ref{fig:long_tail_corr}, a distinct negative correlation is observed.
UDCM exhibits a clear tendency to assign higher confidence to expert predictions for categories with fewer samples, which aligns perfectly with our design motivation to compensate for data scarcity using generalized priors. Additionally, we observed that the global average of $\lambda$ in this model (i.e., use CGL as backbone) exceeds 1.0.
To investigate the rationale behind this distribution, we evaluated the classification accuracy of the two branches.
Surprisingly, the Expert branch processed by LPKEM achieves higher accuracy than the Local Backbone ($84.81\% > 81.47\%$).
This finding strongly validates the effectiveness of our LPKEM: by generating precise masks to filter out background noise, it ensures the expert focuses on intrinsic botanical features, thereby providing a superior predictor than the local model in many cases.

Finally, although the overall weight distribution statistically favors the expert, UDCM is not static; it dynamically adapts to instance-level uncertainty.
We visualize two contrasting decision scenarios in Fig.\ref{fig:case_study}.
In the first scenario (Fig.\ref{fig:case_study}-(a)), the local backbone exhibits high uncertainty in its decision-making.
Detecting this ambiguity, UDCM assigns a dominant weight ($\lambda=1.98$) to the expert, effectively leveraging the expert's prior to rectify the backbone's error.
Conversely, in the second scenario (Fig.\ref{fig:case_study}-(b)), the domain expert provides an incorrect prediction (potentially due to a domain gap), whereas the local backbone makes a correct and high-confidence decision.
Crucially, UDCM successfully identifies this situation and suppresses the calibration weight to a minimal level ($\lambda=0.12$).
This effectively disregards the noisy advice from the expert and preserves the correct local decision.
In summary, these comparisons demonstrate that UDCM can distinguish between when to trust experts and when to rely on local backbones, thus improving classification accuracy.



\subsection{Parameter Sensitivity Analysis}
We analyzed the sensitivity of our framework to the hyperparameter $N$ (i.e., the number of Bin Centers) using the CGL backbone on the CU-Tree102 dataset.
We varied $N$ from 2 to 32 and reported the corresponding classification metrics. As illustrated in Fig. \ref{fig:k_sensitivity}, both Top-1 Accuracy and Macro F1-score achieve their best performance at \textbf{$N=8$}.
When $N$ is too small (e.g., 2) or excessively large (e.g., 32), the model suffers from slight performance degradation due to under-optimization or over-optimization, respectively.
However, the overall performance remains within a highly stable range, with Top-1 Accuracy consistently exceeding \textbf{86\%} across all settings.
This stable performance demonstrates the robustness of EKDC-Net to this hyperparameter.
\section{Conclusion}
\label{sec:conclusion}
This work identifies and addresses a critical limitation in existing tree species classification methods: the overfitting to fixed local distributions, which compromises performance on few-shot and confusing categories. To mitigate this, we introduced EKDC-Net, a framework that integrates an external domain expert. It comprises two specialized modules, LPKEM and UDCM, which are tailored to extract and leverage expert knowledge, respectively, thereby rectifying the biased predictions of the local model. Beyond these methodological innovations, we also released a classification benchmark dataset containing 102 tree species to address the issue of scarce species diversity in current benchmarks.

Our experimental results validated that EKDC-Net not only outperforms current state-of-the-art methods but also maintains high parameter efficiency. By effectively combining the discriminative power of expert knowledge with the local backbones, our work provides a practical and scalable tool for precise forest inventory and city-scale ecological management.

\section*{Acknowledgment}
This work was supported by the National Key Research and Development Program of China under Grant 2023YFF0725200.

\section*{Declaration of generative AI in scientific writing}
During the preparation of this work the author(s) used Gemini in order to polish the manuscript. After using this tool/service, the author(s) reviewed and edited the content as needed and take(s) full responsibility for the content of the publication. Responsibility for the content of the publication.

\section*{Disclosure statement}
The authors affirm that there are no apparent conflicting financial interests or personal connections that could be considered to have swayed the results of the research presented in this publication.

\section*{Author contributions}
Conceptualization: C.L.; Data curation: C.L.,D.C.,R.D.; Methodology:C.L.,Z.C.; Writing-original draft: C.L.; Funding acquisition: Z.D.; Project administration: Z.D.,B.Y.; All authors have read and agreed to the published version of the manuscript.







\bibliographystyle{cas-model2-names}

\bibliography{cas-refs}

\end{document}